\documentclass{article} 
\usepackage{iclr2027_conference,times}

\usepackage{amsmath,amsfonts,bm}

\def\eqref#1{equation~\ref{#1}}

\def\1{\bm{1}}

\def\btheta{{\bm{\theta}}}

\def\vx{{\bm{x}}}

\def\mA{{\bm{A}}}
\def\mB{{\bm{B}}}
\def\mC{{\bm{C}}}

\def\mV{{\bm{V}}}
\def\mW{{\bm{W}}}
\def\mX{{\bm{X}}}

\DeclareMathAlphabet{\mathsfit}{\encodingdefault}{\sfdefault}{m}{sl}
\SetMathAlphabet{\mathsfit}{bold}{\encodingdefault}{\sfdefault}{bx}{n}

\def\gH{{\mathcal{H}}}

\def\gS{{\mathcal{S}}}

\newcommand{\E}{\mathbb{E}}

\newcommand{\R}{\mathbb{R}}

\newcommand{\Erea}{\mathbb E_{\chi\sim\mathcal D_{\rm rea}}}

\newcommand{\attn}{\mathrm{attn}}
\DeclareMathOperator{\MASK}{MASK}

\usepackage{hyperref}
\usepackage{url}
\usepackage{physics}
\usepackage{lmodern,amsmath,amssymb,amsthm,mathtools,bm}
\usepackage{graphicx}
\usepackage{cleveref}
\usepackage{booktabs}
\usepackage{wrapfig}
\usepackage{subcaption}
\usepackage{xcolor}

\title{Understanding the Role of Reasoning Data in Pretraining}

\author{Jianhao Huang$^{1}$\quad  Allan Zhang$^{1}$\quad  Ali Payani$^{2}$\quad  Baharan Mirzasoleiman$^{1}$ \\
$^{1}$University of California, Los Angeles \quad $^{2}$Cisco Research \\
\texttt{jhhuang2025@cs.ucla.edu}}

\newtheorem{theorem}{Theorem}
\newtheorem{lemma}{Lemma}

\newtheorem{definition}{Definition}

\newtheorem{assumption}{Assumption}

\newcommand{\rebuttal}[1]{{#1}}

\iclrfinalcopy 
\begin{document}

\maketitle
\lhead{Preprint}

\begin{abstract}
Reasoning data expose models to multi-step reasoning, but \rebuttal{reasoning traces may cover only a subset of the available facts.} How should pretraining balance atomic facts with reasoning traces? We study this question by constructing a setting in which every fact provides an atomic training signal, while \rebuttal{only a subset appears in reasoning supervision, enabling us to measure generalization to transitions observed atomically but never used in training traces.} For a linear transformer, we derive the weight assigned to atomic training signals that minimizes held-transition squared error and show that its leading-order optimum increases with reasoning-chain length and decreases with the fraction of facts held out from reasoning traces. \rebuttal{These trends persist in a full transformer layer with an MLP, softmax attention, and cross-entropy loss. We further validate the role of reasoning complexity and data coverage in two natural-language settings: controlled multi-hop reasoning and mathematical pretraining.} Together, our results provide a principled view of how reasoning-chain structure \rebuttal{and data coverage} should shape the allocation of training data during pretraining.

\end{abstract}

\section{Introduction}
Reasoning traces teach models how to compose facts across multiple steps, while atomic examples provide broader coverage of those facts. Under a fixed training budget, these objectives compete: spending more data on reasoning traces leaves less budget for learning individual facts, while spending more on atomic examples provides broader factual coverage but less practice with composition. This raises a fundamental question for pretraining: \textbf{how should a model's training budget be divided between atomic facts and reasoning traces?}

This question becomes especially important as reasoning data move into pretraining, rather than being reserved for the later post-training stage. Recent work has begun to examine how reasoning data should be incorporated into pretraining. \citet{akter2026front} study the effects of when reasoning data are introduced during training, as well as their diversity and quality, while \citet{baek2026the} investigate repeatedly sampled domain data during pretraining. Recent training recipes likewise incorporate instruction data and thinking traces into Olmo~3 midtraining~\citep{olmo2025olmo} and SFT-style examples into Nemotron~3 Nano pretraining~\citep{blakeman2025nemotron}. Together, these developments establish reasoning-oriented data as an increasingly important component of the pretraining mixture. Yet they leave open a basic question: \textbf{how should the optimal mixture depend on the structure and coverage of reasoning data?}

The tension is subtle because reasoning traces provide something atomic examples do not: they teach the model how to use knowledge across multiple steps. But reasoning traces can also be an expensive way to teach that knowledge. Longer traces consume more of a fixed training budget, while a limited collection of traces may cover only a subset of the facts needed at evaluation. Thus, longer reasoning can simultaneously make compositional training more valuable and make broad factual coverage more important. This suggests that the optimal allocation should depend on both \textbf{reasoning-chain length} and \textbf{reasoning-trace coverage}, namely, the fraction of facts that appear inside reasoning traces. Our goal is to characterize this allocation under a fixed training budget, where differences in learning dynamics matter even when different data mixtures might eventually reach similar performance.

We study this question first in \textbf{state tracking}, a controlled setting for multi-step reasoning over a finite environment~\citep{merrill2024the, merrill2025a,forner2026learning,huang2026transformers,li2025language}. Starting from an initial state, the model receives a sequence of relations and must infer the state reached after executing the entire sequence. This formulation captures capabilities that arise in practical language-model tasks, including tracking entities through a narrative, reasoning over sequences of chess moves, and executing programs~\citep{kim2023entity,merrill2024the}. It has consequently become a standard testbed for studying language-model reasoning, both theoretically~\citep{liu2023transformers,merrill2024the,merrill2025a} and empirically~\citep{zhang2022unveiling,liu2023transformers,li2025language}.

Our construction makes the distinction between \textbf{learning facts} and \textbf{learning to compose them} explicit. Each state transition corresponds to an atomic fact. We include every transition as an atomic training example, but construct reasoning traces using only a subset of the transitions. The remaining transitions are \emph{held transitions}: the model sees them individually during training, but never encounters them inside a reasoning trace. Thus, a held transition is not an unseen fact---the model has learned it atomically---but an unseen \emph{use} of that fact within a reasoning chain. This construction cleanly separates factual coverage from reasoning composition and lets us ask whether learning to compose facts transfers to facts that have never been used in a reasoning trace.

We first explore this tradeoff empirically using a one-layer transformer with softmax attention and an MLP. We vary reasoning-chain length and reasoning-trace coverage. \textbf{Perhaps surprisingly, we find that longer reasoning chains favor allocating more training data to atomic facts, not less.} Intuitively, longer traces provide richer compositional training, but each trace also consumes more of the fixed budget, making broad factual coverage increasingly important. \rebuttal{Broader reasoning-trace coverage likewise favors a larger allocation to atomic facts.} \rebuttal{We further confirm that the preferred mixture depends on both reasoning complexity and data coverage in two natural-language settings: controlled multi-hop reasoning and pretraining on FineMath~\citep{allal2025smollm} and MathLake~\citep{cai2025opendataarena}.}

To explain these trends, we turn to a linear transformer, a standard simplification for analyzing transformer learning dynamics~\citep{zhang2024trained,ahn2023transformers,huang2025transformers}. Atomic training examples directly reinforce transitions that reasoning traces may never contain, while reasoning traces train the model to compose transitions across multiple steps. We derive the atomic-data weight that minimizes squared prediction error on held transitions. To leading order, the optimal weight \textbf{increases with reasoning-chain length and decreases with the fraction of held transitions}. \rebuttal{Thus, both longer reasoning chains and broader reasoning-trace coverage favor a larger atomic allocation in the analyzed regime.}

\paragraph{Contributions.}
Our contributions are three-fold. First, we introduce a state-tracking setting that separates \textbf{factual coverage} from \textbf{reasoning composition}: all transitions are observed atomically, but only a subset appears inside reasoning traces. Using a one-layer softmax-MLP transformer, we show that an interior mixture of atomic and reasoning examples can outperform either extreme, and characterize how the preferred mixture changes with reasoning-chain length and reasoning-trace coverage. Second, for a linear-attention model, we derive an explicit mixture law whose optimum depends on reasoning-chain length and coverage, \rebuttal{showing how both longer reasoning chains and broader reasoning-trace coverage can favor greater allocation to atomic facts.} \rebuttal{Third, we extend the empirical study to two natural-language settings: a controlled multi-hop task that preserves the separation between atomic facts and reasoning supervision, and a fixed-token FineMath--MathLake pretraining protocol that varies reasoning-data availability, difficulty, and introduction time.}

\section{Related Work}
\paragraph{Reasoning data in pretraining.}
Recent work has increasingly explored the role of reasoning data during pretraining, rather than reserving them for post-training. \citet{akter2026front} systematically study how reasoning data should be allocated between pretraining and supervised fine-tuning, showing that reasoning examples introduced during pretraining provide benefits that cannot be fully recovered by shifting the same supervision to post-training. \citet{wang2025octothinker} find that mathematical and long-chain-of-thought data during mid-training improve both base-model reasoning and subsequent reinforcement-learning performance, while \citet{hatamizadeh2026rlp} incorporate explicit reasoning directly into the pretraining objective. An earlier related observation comes from \citet{allen-zhu2024physics}, who, although not studying reasoning data specifically, argue from controlled knowledge-learning experiments that instruction-finetuning data should be mixed into pretraining rather than reserved entirely for post-training. These works establish the value of introducing richer forms of supervision during pretraining, but leave open a more fundamental question: \emph{how should reasoning data be balanced against atomic data when the training budget is fixed?} We study this tradeoff by separating factual coverage from reasoning composition: every fact is observed atomically, but only a subset is exposed within reasoning traces. This allows us to characterize when reasoning data are most valuable and how their benefit depends on reasoning-chain length and coverage.

\paragraph{Transformer theory.}
A growing body of work uses controlled synthetic tasks to understand how Transformers acquire, retrieve, and compose structured knowledge. \citet{nichani2025understanding} analyze factual recall through associative memories, showing how shallow Transformers can store and retrieve large collections of subject-relation-object associations. \citet{huang2026fine} study how attention learns to select relevant relations and retrieve corresponding facts from a pretrained associative memory. Most closely related to our setting, \citet{huang2026transformers} prove that Transformers trained by gradient descent can learn chain-of-thought state tracking and characterize how the algebraic structure of the transition system governs length generalization through attention concentration. These works provide insight into how Transformers store and compose knowledge, but do not study how different forms of training data interact under a fixed budget. We instead consider a setting in which every transition is observed through atomic examples, while only a subset appears in reasoning traces. This cleanly separates \textbf{factual availability} from \textbf{exposure to multi-step composition}, allowing us to analyze how the allocation between atomic and reasoning data affects the ability to compose transitions that are known individually but absent from the reasoning data.

\section{Toy setting: state tracking}
\subsection{Preliminaries}
\begin{definition}[State-tracking world and held transitions]
\label{def:task}
Let $[K]=\{1,\ldots,K\}$ and $N,M\ge2$. Entities are indexed by
$e\in[N]$, and each relation $r\in[M]$ acts through a fixed permutation
$\pi_r:[N]\to[N]$. The full fact table
$\mathcal A=[N]\times[M]$ contains $D=NM$ entity--relation pairs.
A length-$L$ chain
$\chi=(r_1,\ldots,r_L,e_0)\in[M]^L\times[N]$, with $L\ge2$,
induces the state sequence
\begin{equation*}
 e_\ell=\pi_{r_\ell}(e_{\ell-1}),
 \qquad \ell\in[L].
\end{equation*}
For an integer $1\le m<D$, draw the held set $\gH$ and define its
complement $\gS$ by
\begin{equation*}
 \gH\sim\operatorname{Unif}
       \{H\subseteq\mathcal A:|H|=m\},
 \qquad
 q=\frac{m}{D},\quad p=1-q,\quad
 \gS=\mathcal A\setminus\gH.
\end{equation*}
The \emph{held fraction} $q$ is the fraction of facts excluded from
reasoning traces, and the \emph{reasoning coverage} $p$ is the fraction
available to those traces. The mask is sampled once and fixed
throughout each run.
\end{definition}

\begin{definition}[Atomic, reasoning, and evaluation populations]
\label{def:populations}
An atomic example samples $e_0,r_1,r_2,\cdots, r_L$ independently and
uniformly, but supervises only $e_1=\pi_{r_1}(e_0)$. A reasoning example is sampled uniformly
from chains whose actual transitions all belong to $\gS$ and
supervises the complete state trace $e_1,\ldots,e_L$. Formally,
\begin{align*}
\mathcal D_{\rm at}=\mathcal D_{\rm full}
&=\operatorname{Unif}([M]^L\times[N]),\\
\mathcal C_L(\gS)
&=\bigl\{\chi\in[M]^L\times[N]:
   (e_{\ell-1},r_\ell)\in\gS
   \text{ for every }\ell\in[L]\bigr\},\\
\mathcal D_{\rm rea}(\gS,L)
&=\operatorname{Unif}(\mathcal C_L(\gS)).
\end{align*}
\end{definition}

We write $\Erea$ for expectation over the reasoning population
associated with the fixed set $\gS$. 

\paragraph{Learning objective.}
Entities and relations are represented by distinct typed identity
tokens. Both data sources use the same relation-first input format:
\begin{equation*}
 \begin{array}{rcl}
 \text{atomic:}
   &[r_1,\cdots,r_L,e_0]&\longrightarrow[e_1],\\
 \text{reasoning:}
   &[r_1,\ldots,r_L,e_0]&\longrightarrow[e_1,\ldots,e_L].
 \end{array}
\end{equation*}
The shared prefix places the first prediction at the same absolute
position and supplies the same number of candidate relations in both
sources. Atomic examples supervise only the first transition, whereas
reasoning examples additionally supervise subsequent prediction roles.
Every fact is available through atomic supervision, while reasoning
traces cover only a subset, allowing us to study whether facts learned
individually can be used within multi-step computations.

Let $\beta\in[0,1]$ be the atomic \emph{loss weight}. We average the
reasoning loss over its $L$ state predictions before combining the
two sources:
\begin{equation}\label{eq:ce}
 \begin{aligned}
 \mathcal L_\beta^{\rm CE}(\theta)
 ={}&\beta\,\E_{\chi\sim\mathcal D_{\rm at}}
        \ell_{{\rm CE},1}(\theta;\chi)\\
 &+(1-\beta)\,\Erea\left[
      \frac1L\sum_{\ell=1}^{L}
      \ell_{{\rm CE},\ell}(\theta;\chi)
    \right],
 \end{aligned}
\end{equation}
where
\begin{equation*}
 \ell_{{\rm CE},\ell}(\theta;\chi)
 =-\log p_\theta\left(
     e_\ell\mid[r_1,\ldots,r_L,e_0,\ldots,e_{\ell-1}]
   \right),
 \qquad \ell\in[L].
\end{equation*}
Each cross-entropy term supervises the next entity using the true
preceding states. A test chain is called \emph{held} if at least one of its
ground-truth transitions belongs to $\gH$.

\subsection{One full layer transformer}
Our learner consists of a single causal self-attention head followed by a position-wise multilayer perceptron (MLP). We encode token identities and positions in separate one-hot blocks and directly parameterize the merged query-key matrix, following \citet{zhang2024trained, ahn2023transformers,huang2025transformers,huang2026transformers,huang2026fine}. We first define the input
representation, followed by the attention and output maps.

\begin{definition}[Token and positional encoding]

Let $u_r\in\R^M$ and $v_e\in\R^N$ be the standard basis vectors
corresponding to relation $r$ and entity $e$, respectively. Fix a
maximum input-prefix length $T\ge2L$, shared by atomic and reasoning
examples. The token-content vectors are
\begin{equation*}
 c(r)=\begin{pmatrix}u_r\\0_N\end{pmatrix},
 \qquad
 c(e)=\begin{pmatrix}0_M\\v_e\end{pmatrix}.
\end{equation*}
For a token $x_t$ at position $t$, let
$\boldsymbol c_t:=c(x_t)$ and let $\boldsymbol p_t\in\R^T$ be the
$t$-th standard basis vector. Concatenating content and position gives
\begin{equation*}
 \widetilde{\vx}_t
 =\begin{pmatrix}\boldsymbol c_t\\\boldsymbol p_t\end{pmatrix}
 \in\R^d,
 \qquad
 d=M+N+T,
 \qquad
 \widetilde{\mX}_t
 =[\widetilde{\vx}_1,\ldots,\widetilde{\vx}_t]
 \in\R^{d\times t}.
\end{equation*}
All content and positional vectors are fixed. Relation, entity, and
position coordinates occupy disjoint blocks of sizes $M$, $N$, and
$T$, respectively. We write
$\mX_t=[\boldsymbol c_1,\ldots,\boldsymbol c_t]
\in\R^{(M+N)\times t}$ for the corresponding content-only matrix.
\end{definition}

Both data sources predict $e_1$ at position $L+1$, and reasoning
role $\ell$ predicts $e_\ell$ from the prefix ending at position
$L+\ell$. The last supervised query is therefore at position $2L$,
so $T\ge2L$ covers all input prefixes. We reserve $T$ for the maximum
prefix length and use $\tau$ for the training horizon below.

We parameterize the merged query--key scores directly by a trainable
matrix $\mA\in\R^{d\times d}$. In particular, the score assigned to
key position $i$ by query position $j$ is
$\widetilde{\vx}_i^\top\mA\widetilde{\vx}_j$.

\begin{definition}[Causal self-attention head]
\label{def:attn}
For $\mA\in\R^{d\times d}$ and a prefix
$\widetilde{\mX}_t\in\R^{d\times t}$, define
\begin{equation*}
 \attn(\widetilde{\mX}_t;\mA)
 :=
 \widetilde{\mX}_t
 \operatorname{softmax}\left(
   \MASK(\widetilde{\mX}_t^\top\mA\widetilde{\mX}_t)
 \right)
 \in\R^{d\times t},
\end{equation*}
where, for $S\in\R^{t\times t}$,
\[
 [\MASK(S)]_{ij}
 =
 \begin{cases}
  S_{ij}, & i\le j,\\
  -\infty, & i>j.
 \end{cases}
\]
The softmax is applied column-wise.
\end{definition}

The attention output is passed through a shared MLP to produce
next-token logits over the entity and relation vocabulary.

\begin{definition}[One-layer attention--MLP Transformer]
\label{def:transformer}
Fix an MLP width $m\ge1$ and a coordinatewise
nonlinearity $\phi$. The trainable matrices are
\[
 \mA\in\R^{d\times d},
 \qquad
 \mW\in\R^{m\times d},
 \qquad
 \mV\in\R^{m\times(M+N)},
\]
with parameters $\btheta_{\mathrm{TF}}=(\mA,\mV,\mW)$.
At query position $t$, the vocabulary-logit vector is
\begin{equation*}
 f_{\mathrm{TF},t}
   (\widetilde{\mX}_t;\btheta_{\mathrm{TF}})
 =
 \mV^\top\phi\left(
   \mW[\attn(\widetilde{\mX}_t;\mA)]_{:,t}
 \right)
 \in\R^{M+N}.
\end{equation*}
The next-token distribution is
\begin{equation*}
 p_{\btheta_{\mathrm{TF}}}
   (\cdot\mid\widetilde{\mX}_t)
 =
 \operatorname{softmax}\left(
   f_{\mathrm{TF},t}
     (\widetilde{\mX}_t;\btheta_{\mathrm{TF}})
 \right).
\end{equation*}
The first $M$ output coordinates index relations, and the last $N$
index entities. Accordingly, the probability assigned to entity
$e\in[N]$ is component $M+e$ of this distribution.
\end{definition}

We train this model using the mixed cross-entropy objective
in~\eqref{eq:ce}.

\begin{figure}[t]
    \centering
    \includegraphics[width=\linewidth]{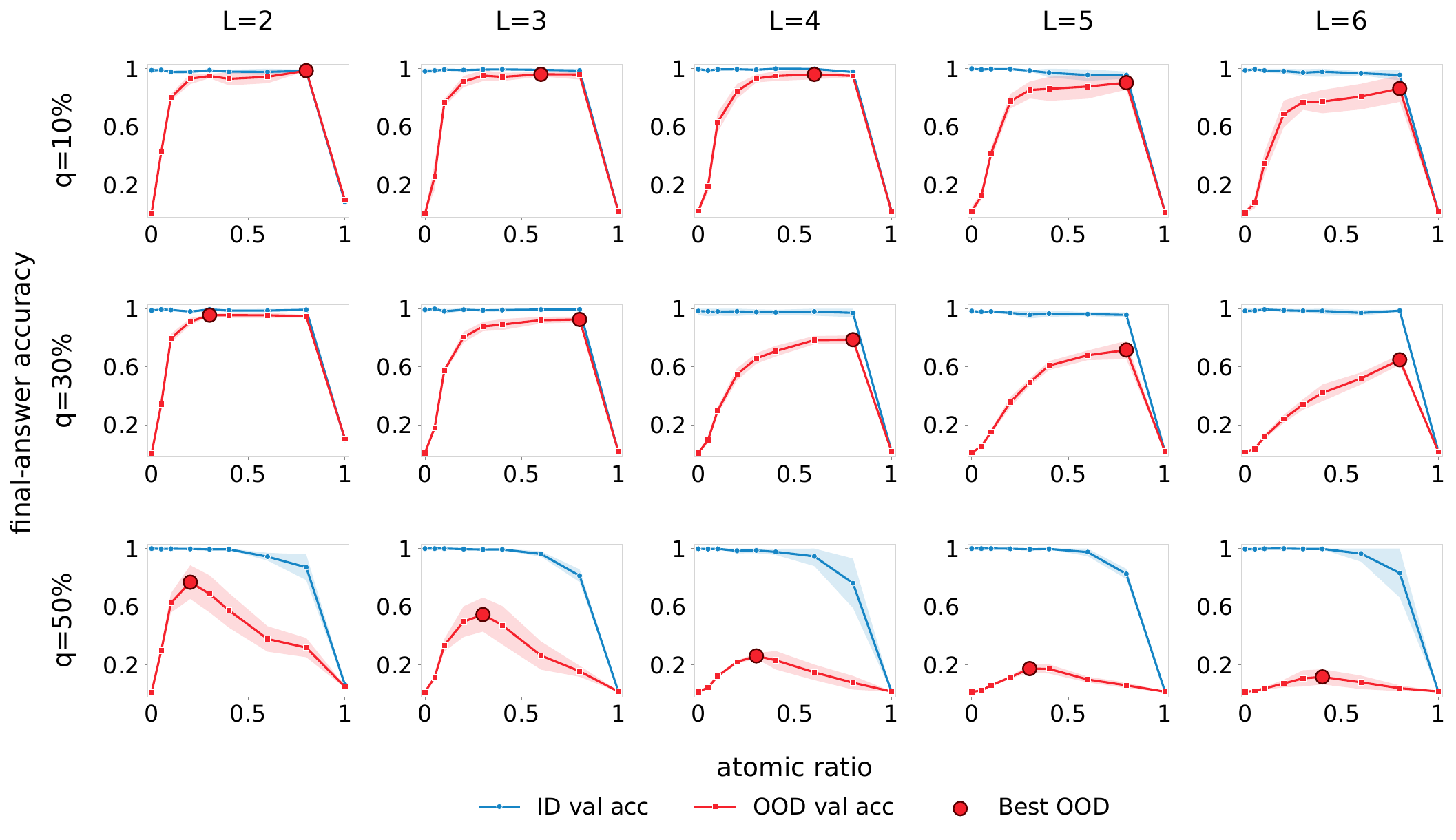}
    \caption{\textbf{Effect of atomic--reasoning mixing on state-tracking accuracy.}
Final-answer accuracy as a function of the atomic ratio in a
synthetic world with $64$ entities and $8$ relations, each sampled as an
independent random permutation. 
Curves and shaded regions indicate the mean and $\pm1$ standard
deviation across three training seeds.
The empirically optimal atomic example ratio generally increases with
reasoning length $L$ and decreases with the held-out fact fraction $q$.
The selected grid maximizers for $L\in\{3,4\}$ at $q=30\%$ deviate
from the latter trend; however, these panels exhibit near-optimal
plateaus, with several tested mixture ratios achieving very similar
mean OOD accuracy.}
    \label{fig:toy}
\end{figure}

\paragraph{Effect of data mixture}
Figure~\ref{fig:toy} shows that mixing atomic and
reasoning examples improves OOD accuracy over either pure-source
baseline, suggesting that the two forms of supervision play
complementary roles. The empirically preferred atomic example ratio
generally increases with reasoning length $L$ and decreases with the
held-out fact fraction $q$. This pattern is not strictly monotone across
all selected grid maximizers: the $L\in\{3,4\}$ cases at $q=30\%$ deviate
from the held-fraction trend. However, these panels exhibit near-optimal
plateaus, with several mixture ratios attaining very similar mean OOD
accuracy.  We refer to \Cref{app:toy} for more experiment details.

\subsection{Linear-attention and the optimal mixture}\label{sec:lineartf}
We next examine how the allocation between atomic examples and reasoning
traces affects optimization. Our goal is to characterize the mixture that
best supports the use of held transitions after a common training horizon,
and to understand how this allocation depends on reasoning length and
coverage. To make this interaction tractable, we study a scalar
linear-attention surrogate

Following prior block-structured analyses of Transformer learning
\citep{nichani2024how,huang2025transformers,huang2026how,huang2026fine,javanmard2026understanding,javanmard2026theoretical}, we use a block-diagonal attention
score matrix with token-token and position-position blocks,
fixing the mixed token-position blocks to zero. These two components
contribute additively to the softmax logits, but their contributions
multiply after exponentiation. This motivates combining the bilinear
content scores with a learned positional gate through the Hadamard
product in the linear surrogate:
\begin{equation*}
 (\mX_t^\top\mB\mX_t)\odot\widetilde{\mC}_{1:t,\,1:t}.
\end{equation*}
Here $\mB$ controls token interactions, while $\widetilde{\mC}$ controls
positional routing. In particular, a zero gate entry excludes a
connection, whereas an additive zero would leave its content score
unchanged. Removing attention normalization and replacing the nonlinear
output map with a scalar linear readout gives the following model.

\begin{definition}[Scalar linear-attention model]
\label{rd23:def:linattention}
Let $\mX_t\in\R^{(M+N)\times t}$ contain the fixed typed one-hot content
vectors, with $t\le T$ and $T\ge2L$. For
$\mW\in\R^{1\times(M+N)}$,
$\mB\in\R^{(M+N)\times(M+N)}$, and
$\widetilde{\mC}\in\R^{T\times T}$, define
\begin{equation}\label{rd23:eq:linhead}
 f_{\mathrm{lin}}(\mX_t;\mW,\mB,\widetilde{\mC})
 =\mW\mX_t\left[
       (\mX_t^\top\mB\mX_t)\odot
       \widetilde{\mC}_{1:t,\,1:t}
     \right]
 \in\R^{1\times t}.
\end{equation}
The trainable coordinates are the relation values
$w_r=\mW_{1,r}$, the relation--entity scores
$b_{r,e}=\mB_{r,M+e}$, and the routing coefficients
$C_{j\ell}=\widetilde{\mC}_{j,L+\ell}$ for $j,\ell\in[L]$;
all remaining entries are fixed at zero. Write $\theta=(w,b,C)$.

At role $\ell$, the teacher-forced prefix ends at $e_{\ell-1}$ and has
length $L+\ell$. Let $f_\ell(\chi;\theta)$ be entry $(1,L+\ell)$ of
$f_{\mathrm{lin}}(\mX_{L+\ell};\theta)$. For entity $e\in[N]$, define
\begin{equation}\label{rd23:eq:scalarcode}
 \zeta_e=\frac{2e-N-1}{N},
 \qquad
 \widehat e_\ell\in\operatorname*{arg\,min}_{e\in[N]}
       \left|f_\ell(\chi;\theta)-\zeta_e\right|.
\end{equation}
\end{definition}




\paragraph{Squared loss and joint training.}
Atomic prefixes sample the initial
entity and all relation labels independently and uniformly; reasoning
prefixes are uniform over chains whose actual transitions belong to
$\gS$. For an atomic loss weight $\beta\in[0,1]$, the mixed population
loss is
\begin{equation}\label{rd23:eq:objective}
 \begin{aligned}
 L_\beta(\theta;\gS)
 ={}&\frac\beta2\E_{\chi\sim\mathcal D_{\mathrm{at}}}
       \left(f_1(\chi;\theta)-\zeta_{e_1}\right)^2\\
 &+\frac{1-\beta}{2L}\sum_{\ell=1}^L
       \Erea\left(f_\ell(\chi;\theta)-\zeta_{e_\ell}\right)^2.
 \end{aligned}
\end{equation}
Atomic examples supervise only the first state, whereas reasoning
examples supervise the complete trace. Averaging the reasoning loss over
its $L$ predictions makes $\beta$ the relative weight of the two source
losses.

All active entries of $\mW,\mB,\widetilde{\mC}$ are jointly trained by gradient flow:
\begin{equation}\label{rd23:eq:flow}
 \dot\theta_\beta(t)
 =-\nabla_\theta L_\beta(\theta_\beta(t);\gS),
 \qquad \theta_\beta(0)=\theta_0.
\end{equation}
We initialize $\mB$ to zero and use the same initialization for every
mixture:
\begin{equation}\label{rd23:eq:init}
 \mB(0)=0,
 \qquad w_r(0)=1\quad(r\in[M]),
 \qquad C(0)=E_{11}.
\end{equation}
Here $E_{11}$ has a single unit entry at $(1,1)$ and zeros elsewhere.

\paragraph{Held-transition error and optimal allocation.}
The training objective alone does not measure whether facts learned from
atomic supervision can be used at later prediction roles. We therefore
evaluate later transitions that were excluded from reasoning traces.
For each $\ell\ge2$, let
\[
 \mathcal D_{H,\ell}
 :=\mathcal D_{\mathrm{full}}
    \bigl(\,\cdot\mid(e_{\ell-1},r_\ell)\in\gH\bigr).
\]
The conditioning event has probability $q>0$, so this distribution is
well-defined. Define the held-transition error by
\begin{equation*}
 \mathcal E_H(\theta)
 :=\frac1{2(L-1)}\sum_{\ell=2}^L
       \E_{\chi\sim\mathcal D_{H,\ell}}
       \left(f_\ell(\chi;\theta)-\zeta_{e_\ell}\right)^2.
\end{equation*}

At a training horizon $\tau>0$, define the optimal atomic weight as
\begin{equation}\label{rd23:eq:optdefinition}
 \beta_{\tau,S}^{\star}
 \in\operatorname*{arg\,min}_{\beta\in[0,1]}
       \mathcal E_H\left(\theta_\beta(\tau)\right).
\end{equation}

\rebuttal{Motivated by the simply transitive state-tracking setting of
\citet{huang2026transformers}, we impose the following balance
condition on the transition world.

\begin{assumption}[Balanced state-tracking world]
\label{cs:ass:balanced-world}
The world consists of $N\ge2$ entities and $M=N$ relations.
Each relation $r\in[N]$ acts through a fixed permutation
$\pi_r:[N]\to[N]$. Moreover, for every pair of entities
$e,e'\in[N]$, exactly one relation maps $e$ to $e'$:
\begin{equation}\label{cs:eq:world-balance}
    \sum_{r=1}^{N}
    \mathbf{1}\{\pi_r(e)=e'\}=1.
\end{equation}
Equivalently, for each fixed entity $e$, the map
$r\mapsto\pi_r(e)$ is a bijection of $[N]$.
The transition world is fixed throughout training and evaluation.
\end{assumption}

Under \Cref{cs:ass:balanced-world}, a uniformly sampled relation
maps any current entity to a uniformly distributed next entity.
In particular, for the scalar encoding
$\zeta_e=(2e-N-1)/N$, every $e\in[N]$ satisfies
\begin{equation}\label{cs:eq:scalar-balance}
    \frac1N\sum_{r=1}^{N}\zeta_{\pi_r(e)}=0,
    \qquad
    \frac1N\sum_{r=1}^{N}\zeta_{\pi_r(e)}^2
    =\frac{N^2-1}{3N^2}.
\end{equation}
The complete cyclic-shift family
$\pi_r(e)=1+((e+r-2)\bmod N)$ satisfies this assumption.}

We now characterize how reasoning length and coverage affect the optimal
atomic loss weight. The following theorem shows that, at a fixed training
horizon, $\beta^*_{\tau,S}$ concentrates around a deterministic center as the numbers of entities and relations grow.
This center increases with reasoning length and decreases with the held
fraction, describing how the preferred allocation changes with the task
and the available reasoning supervision.

\begin{theorem}[Length--coverage trend]
\label{cs:thm:main}
Fix $L\ge2$, $p=1-q\in(0,1)$, $\tau>0$, and $\delta\in(0,1)$.
Suppose \Cref{cs:ass:balanced-world} holds.
Consider gradient flow on \Cref{rd23:eq:objective}, initialized at
$B(0)=0$, $w_r(0)=1$, and $C(0)=E_{11}$.
Let $\gH$ be uniform among sets of $qN^2$ facts, where $qN^2$
is an integer, and put $\gS=\mathcal A\setminus\gH$.
For all sufficiently large $N$, with probability at least $1-\delta$, the finite-time MSE-optimal
atomic weight $\beta^*_{\tau,S}$ is unique in $(1/3,2/3)$, and
\begin{equation}\label{cs:eq:bound}
 \left|\beta^*_{\tau,S}-\beta\bigl(L(1-q)\bigr)\right|
 \le \mathfrak R_N,
 \qquad
 \mathfrak R_N=O_{L,p,\tau}\!\left(
        \frac{\sqrt{\log(2/\delta)}}{N}\right),
\end{equation}
where
\begin{equation}\label{cs:eq:center}
 \beta(z)=\frac{1}{2-z+\sqrt{z^2-z+1}},\qquad z>0.
\end{equation}
The center increases with $L$ and decreases with $q$. 
\end{theorem}

We refer the complete proof to \Cref{rd23:app:proof}.
The theorem predicts a larger leading optimal atomic weight for longer reasoning chains and a smaller weight when a larger fraction of facts is excluded from reasoning traces, which is consistent with \Cref{fig:toy}.

\section{Experiments on natural language}

\rebuttal{
\subsection{Controlled multi-hop reasoning}
\label{sec:multihop}

We start with a controlled natural-language task that asks a model to follow $H$ relationships between people and identify the final person, which follows from BIO pretraining in \citet{allen-zhu2024physics}. This setting distinguishes two sources of supervision: atomic fact statements and multi-hop reasoning examples.
We vary the atomic-document fraction $a$ in the training mixture and the fraction $q$ of person--relation pairs withheld from reasoning examples. All facts remain available in the atomic corpus, while OOD queries require at least one pair absent from reasoning supervision.

\begin{figure}[t]
\centering
\includegraphics[width=0.6\linewidth]{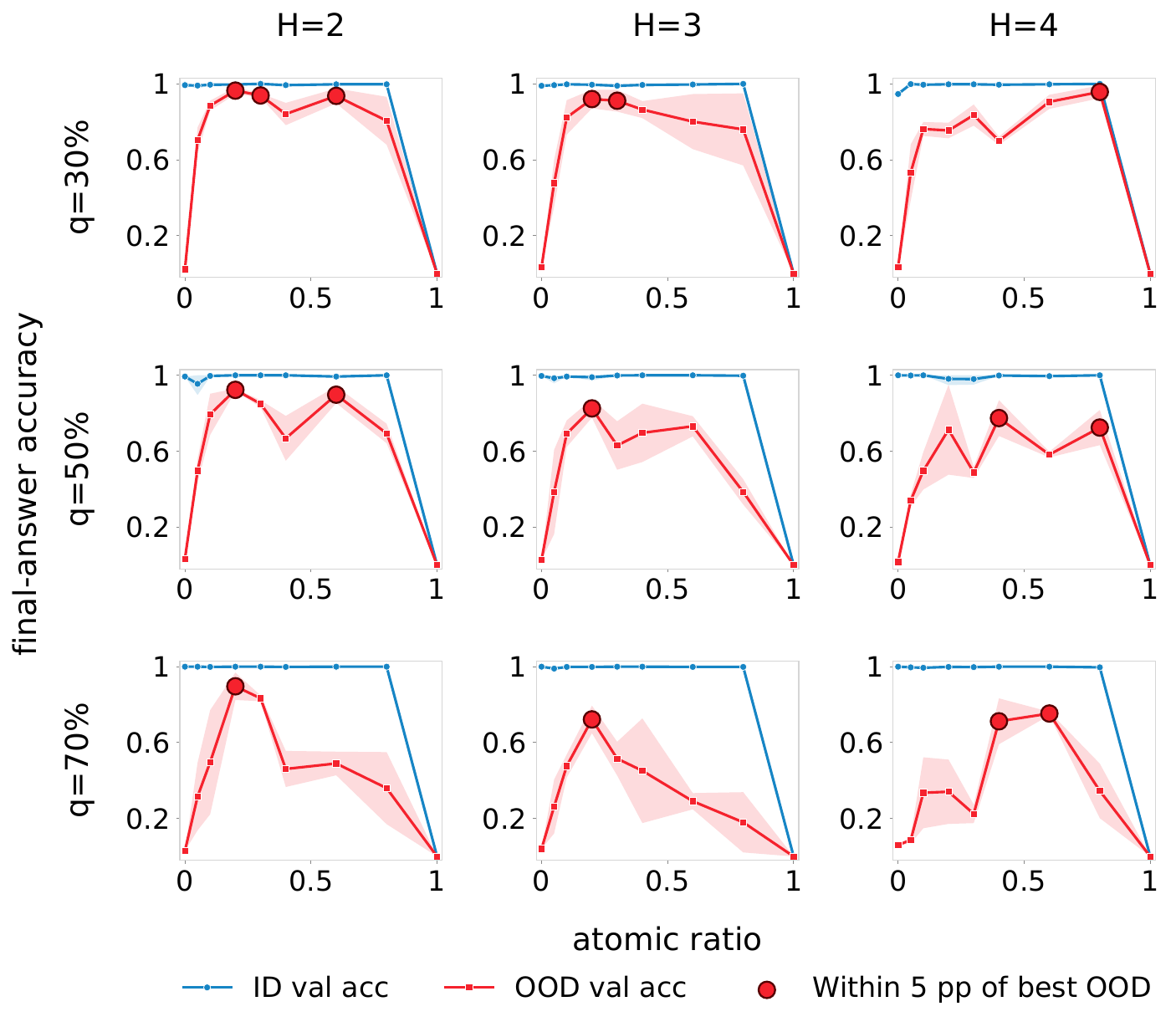}
\caption{\rebuttal{Multi-hop final-answer accuracy as a function of the atomic ratio in a synthetic world with 32 people and 32 relations. Columns vary hop count $H$ and rows vary reasoning holdout fraction $q$. Curves and shaded regions indicate the mean and $\pm1$ standard
deviation across three training seeds. Large red circles mark OOD accuracies within five percentage points of each panel's maximum.}}
\label{fig:multihop-mixture}
\end{figure}

Figure~\ref{fig:multihop-mixture} shows that interior mixtures outperform both pure-source endpoints on OOD accuracy in all displayed settings. Four-hop tasks favor larger atomic fractions than two- and three-hop tasks, while withholding more pairs from reasoning supervision lowers peak OOD accuracy and generally narrows or shifts the near-best region toward smaller atomic fractions. These results are consistent with \Cref{cs:thm:main}. We refer to  Appendix~\ref{app:multihop} for more details. 

\subsection{Pretraining with mathematical reasoning data}
}

\begin{wraptable}{r}{0.4\textwidth}
\vspace{-1.21\baselineskip} 
\centering
\small
\caption{Mixture-ratio sweep between FineMath and MathLake (Olmo-3 150M, 10B tokens). Zero-shot GSM8K test CE and PPL over the solution sequence. Best in bold.}
\label{tab:mixture-sweep}
\vspace{-0.5em}   
\begin{tabular}{rrcc}
\toprule
FM (\%) & ML (\%) & CE $\downarrow$ & PPL $\downarrow$ \\
\midrule
100 & 0 & 1.785 & 5.96 \\
90 & 10 & 1.613 & 5.02 \\
80 & 20 & \textbf{1.608} & \textbf{4.99} \\
70 & 30 & 1.674 & 5.33 \\
60 & 40 & 1.689 & 5.41 \\
50 & 50 & 1.768 & 5.86 \\
40 & 60 & 1.817 & 6.16 \\
30 & 70 & 2.035 & 7.66 \\
20 & 80 & 2.148 & 8.57 \\
10 & 90 & 2.402 & 11.04 \\
0 & 100 & 7.513 & 1831.40 \\
\bottomrule
\end{tabular}
\end{wraptable}

\paragraph{Balancing general math text and reasoning data} 
We pretrain Olmo-3 150M on mixtures of FineMath, a general math web-text corpus, and the OpenMathInstruct-2 subset of MathLake, a mathematical reasoning dataset. We sweep the MathLake fraction from $0\%$ to $100\%$ in $10\%$ steps. The total budget is fixed at 10B tokens, so higher MathLake fractions also mean more repetition of its 277M tokens (up to ${\sim}36$ epochs). Prior work either varies composition and token count together \citep{zhao2025echo}, uses a single reasoning fraction \citep{akter2026front}, or tests only small domain fractions \citep{baek2026the}; none includes both pure-source endpoints, so none shows that a mixture beats either source alone.

Table~\ref{tab:mixture-sweep} shows that both extremes are suboptimal. Pure FineMath reaches a GSM8K CE of 1.785, while pure MathLake degrades to 7.513. Although its training loss falls to 0.2, GSM8K CE rises throughout training, consistent with overfitting to a heavily repeated corpus. A small share of reasoning data helps significantly, with CE dropping to 1.613 at $10\%$ MathLake and reaching its minimum of 1.608 at $20\%$. Beyond this point, CE rises monotonically, returning to the pure FineMath level at $50\%$. We therefore run subsequent experiments at $10\%$, $20\%$, and $30\%$ MathLake, covering the optimum and the start of the degradation regime.

\begin{figure}[h]
\centering

\begin{subfigure}[t]{0.49\linewidth}
  \centering
  \includegraphics[width=\linewidth]{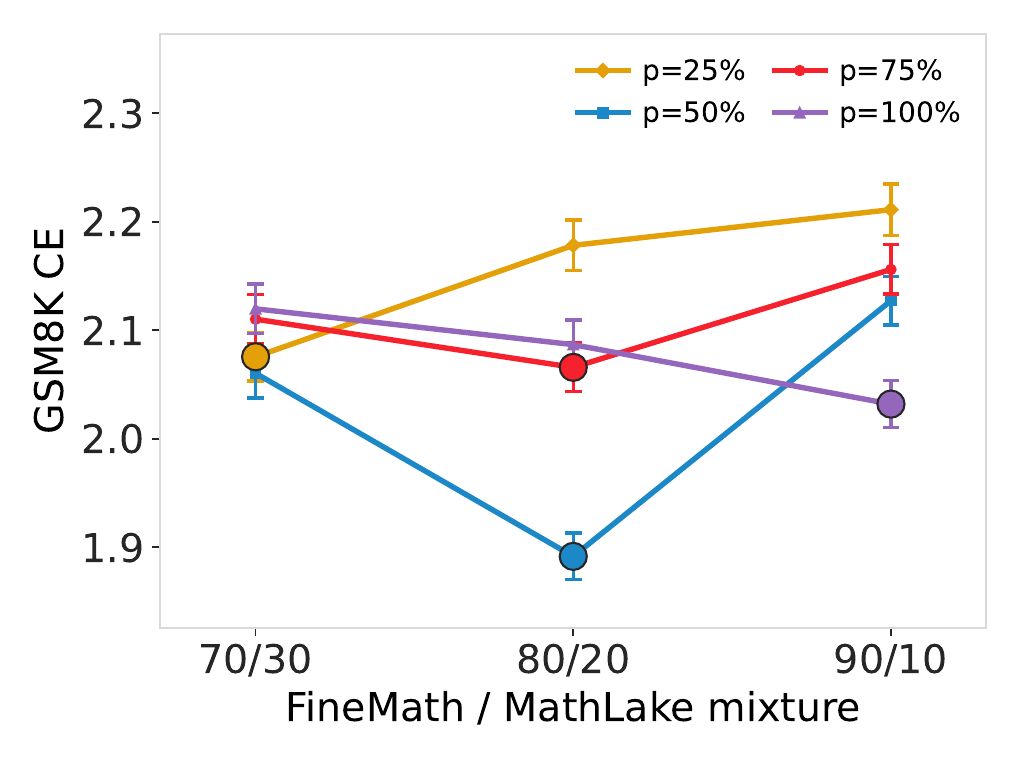}
  \caption{Held-out ratio}
  \label{fig:heldout-ratio}
\end{subfigure}\hfill
\begin{subfigure}[t]{0.49\linewidth}
  \centering
  \includegraphics[width=\linewidth]{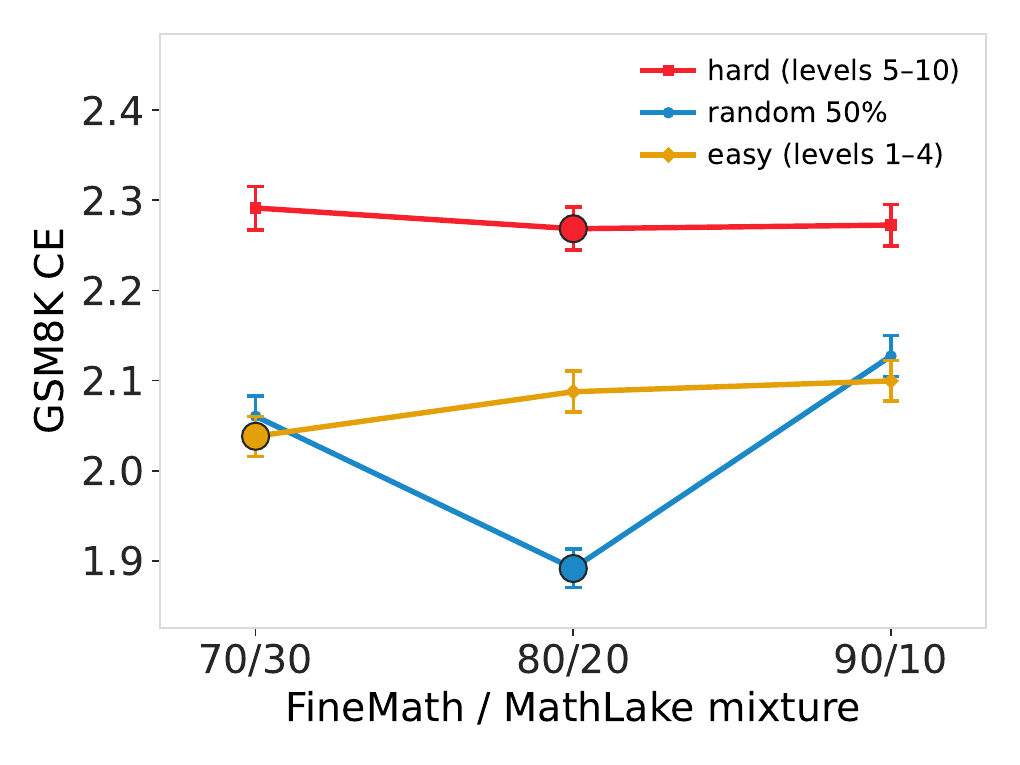}
  \caption{Difficulty (equal-token halves)}
  \label{fig:difficulty}
\end{subfigure}
\caption{GSM8K CE at step 5000, with uncertainty estimated by bootstrapping evaluation examples 3 times. (a) Varying the fraction $p$ of the MathLake pool available for training, at three FineMath/MathLake ratios. (b) Training on the easy or hard half of MathLake versus a random 50\% subset.}
\label{fig:quantity-difficulty}
\end{figure}

\paragraph{Quantity and difficulty of reasoning data} At each of the mixture ratios above, we vary two properties of the reasoning data.

\textit{Held-out ratio.} We control the fraction $p \in \{25\%, 50\%, 75\%, 100\%\}$ of the MathLake pool available for training and withhold the rest. Since the mixture ratio and total budget are fixed, every run sees the same number of reasoning tokens. To make the held-out ratio a monotone intervention, the pools are nested. We shuffle MathLake once at the document level and take each pool as a prefix of the next, where the $25\%$ pool is a subset of the $50\%$ pool, which is a subset of the $75\%$ pool. Thus, moving from a smaller to a larger $p$ only adds unseen documents and never swaps one sample for another.

\textit{Difficulty.} We use MathLake's per-problem difficulty scores (1--10) as a proxy for reasoning length and complexity. The scores closely track solution length (258 tokens at level 1 vs.\ 680 at level 9), so splitting by level alone would also split by token volume. We instead cut the corpus at its token-weighted median, which falls between levels 4 and 5, and cap both halves at 133.9M tokens, so easy and hard runs see identical reasoning-token budgets at identical repetition rates. A random $50\%$ pool (138.7M tokens) serves as the control.

As shown by \Cref{fig:quantity-difficulty}a, the optimal mixing ratio increases with $p$, therefore decreases with held-out ratio. In \Cref{fig:quantity-difficulty}b, the optimal mixing ratio increases with difficulty. The results are consistent with \Cref{cs:thm:main}.

\paragraph{When to introduce reasoning data} Finally, we fix the total reasoning budget and vary when it enters training. Each run lasts 10k steps with MathLake making up $20\%$ of all training tokens, but the first $K$ steps use FineMath only. The full reasoning budget must then fit into the remaining $10\text{k}-K$ steps, so MathLake's share of the post-switch mixture is $20\%$, $25\%$, $33\%$, $50\%$, and $100\%$ for $K = 0, 2\text{k}, 4\text{k}, 6\text{k}, 8\text{k}$. We implement this by resuming each arm from the step-$K$ checkpoint of a single pure-FineMath run, carrying over the step count and learning-rate schedule.

Delaying reasoning data brings no benefit (Table~\ref{tab:timing}): GSM8K CE increases monotonically with $K$. Since starting early is never worse, we recommend mixing in reasoning data from the beginning of training, consistent with \citet{baek2026the}, who find that domain data is best introduced early.

\begin{table}[t]
\centering
\caption{Timing of reasoning data. Each run trains Olmo-3 150M for 10k steps with MathLake making up $20\%$ of all tokens; the first $K$ steps use pure FineMath, so the MathLake share while mixing rises with $K$. Zero-shot GSM8K test CE and PPL over the solution sequence at the final checkpoint. Best in bold.}
\label{tab:timing}
\vspace{4pt}
\begin{tabular}{rccc}
\toprule
Pure FM steps ($K$) & ML share while mixing & CE $\downarrow$ & PPL $\downarrow$ \\
\midrule
0     & 20.0\%  & \textbf{1.744} & \textbf{5.72} \\
1k    & 22.2\%  & 1.749 & 5.75 \\
2k    & 25.0\%  & 1.754 & 5.78 \\
3k    & 28.6\%  & 1.767 & 5.85 \\
4k    & 33.3\%  & 1.771 & 5.88 \\
6k    & 50.0\%  & 1.853 & 6.38 \\
8k    & 100.0\% & 2.411 & 11.14 \\
\bottomrule
\end{tabular}
\end{table}

\section{Conclusion}
We studied a fundamental question in reasoning-oriented pretraining: how should a fixed training budget be divided between atomic facts and reasoning traces? By separating factual coverage from exposure to multi-step composition, we showed that the optimal allocation depends systematically on reasoning-chain length and reasoning-trace coverage. Perhaps surprisingly, longer reasoning chains can favor greater allocation to atomic facts, as richer compositional training comes at the cost of broader factual coverage. Our theoretical analysis explains this effect in a simplified linear setting, and experiments with full Transformers and realistic mathematical data demonstrate that the same qualitative trends persist beyond the controlled regime. Together, our results provide a principled perspective on how the structure and coverage of reasoning data should shape pretraining mixtures. \looseness=-1

\paragraph{Limitations.}
Our controlled setting abstracts away heterogeneous data sources and varying data quality in large-scale pretraining. We also focus on pretraining and do not study how different data mixtures affect subsequent SFT and RL, which we leave for future work.

\subsection*{Acknowledgments}
This research was supported by NSF CAREER Award 2146492, NSF-Simons AI Institute for Cosmic Origins (CosmicAI), NSF AI Institute for Foundations of Machine Learning (IFML), and an Apple
Research Award.

\subsection*{AI use statement}

We have not used generative AI tools for tasks with required disclosure.
Additionally, we used generative AI tools for polishing the paper. We have reviewed all AI-assisted work.  We take responsibility for the final content of this work,
including text, claims or artifacts produced with the aid of generative AI.



\bibliography{iclr2027_conference}
\bibliographystyle{iclr2027_conference}

\appendix

\section{Proof of the length-coverage theorem}\label{rd23:app:proof}
\providecommand{\csproofstep}[1]{\par\smallskip\noindent\textbf{#1}\enspace\ignorespaces}
Throughout, $y_{r,e}=\zeta_{\pi_r(e)}$, $a_{\ell j}=(e_{\ell-1},r_j)$, $a_\ell=a_{\ell\ell}$, $D=N^2$, $\lambda=1-\beta$, and $B(0)=0$.
For a family $h(\beta)$, define
\begin{equation}\label{cs:eq:norm}
 \|h\|_{[2]}:=\sum_{k=0}^2\frac1{k!}
       \sup_{\beta\in[0,1]}\|\partial_\beta^k h(\beta)\|.
\end{equation}
The base norm is entrywise maximum for tables/vectors and
$\|C\|_1=\sum_{j,\ell}|C_{j\ell}|$. A multilinear map of
base operator norm $c$ satisfies
\begin{align}\label{cs:eq:product}
 \|\mathcal B(h_1,h_2)\|_{[2]}
 &\le c\sum_{i+j\le2}
    \frac{\sup_\beta\|\partial_\beta^i h_1\|}{i!}
    \frac{\sup_\beta\|\partial_\beta^j h_2\|}{j!}
 \le c\|h_1\|_{[2]}\|h_2\|_{[2]},\notag\\
 \|\beta\|_{[2]}&=\|1-\beta\|_{[2]}=2.
\end{align}
The same calculation applies to higher multilinear products.

\begin{lemma}[Balance and the fixed-mask cubic]\label{cs:lem:balance}
For a fixed mask with a nonempty reasoning population, put
\begin{equation}\label{cs:eq:moments}
 \begin{gathered}
 \nu=\frac{N^2-1}{3N^2},\qquad
 \nu_H=\frac1{qD}\sum_{a\in H}y_a^2,\qquad
 \omega_a=\frac DL\Pr_S(a_1=a),\\
 K_{j\ell}=\mathbb E_S[y_{a_{\ell j}}y_{a_\ell}],\qquad
 U_{j\ell}=\mathbb E_S[\omega_{a_{\ell j}}y_{a_{\ell j}}y_{a_\ell}],\\
 Z_S=\frac1{qDN}\sum_e
   \left(\sum_{r:(e,r)\in H}y_{r,e}\right)
   \left(\sum_r\omega_{r,e}y_{r,e}\right).
 \end{gathered}
\end{equation}
For $\ell\ge2$,
\begin{equation}\label{cs:eq:balanced-eval}
 \begin{aligned}
 \mathbb E_{H,\ell}[y_{a_{\ell j}}y_{a_\ell}]
   &=\nu_H\mathbf1\{j=\ell\},\\
 \mathbb E_{H,\ell}[\omega_{a_{\ell j}}y_{a_{\ell j}}y_{a_\ell}]
   &=Z_S\mathbf1\{j\ne\ell\}.
 \end{aligned}
\end{equation}
Define
\begin{equation}\label{cs:eq:PS}
 \begin{aligned}
 P_S(\beta)={}&\frac{\nu_H\beta\lambda}{L-1}
     \sum_{\ell=2}^L[\beta K_{\ell\ell}+\lambda U_{\ell\ell}]\\
 &+\frac{Z_S\lambda^2}{L-1}
     \sum_{\ell=2}^L\sum_{j\ne\ell}
       [\beta K_{j\ell}+\lambda U_{j\ell}].
 \end{aligned}
\end{equation}
This is the leading gain polynomial identified in Lemma~\ref{cs:lem:time}.
\end{lemma}
\begin{proof}
Complete shifts and the scalar code give
\begin{equation}\label{cs:eq:balance-proof}
 \frac1N\sum_r y_{r,e}=\frac1N\sum_{x=1}^N\zeta_x=0,
 \qquad
 \frac1N\sum_r y_{r,e}^2=\frac1N\sum_{x=1}^N\zeta_x^2=\nu,
 \qquad \frac14\le\nu<\frac13.
\end{equation}
For every relation tuple $\boldsymbol r$,
\begin{equation}\label{cs:eq:conditioning-proof}
 \Pr_{\rm full}(e_{\ell-1}=e\mid\boldsymbol r)=\frac1N,
 \quad \Pr_{\rm full}(a_\ell=a)=\frac1D,
 \quad \Pr_{\rm full}(r_j=t\mid a_\ell=(e,r))=\frac1N\quad(j\ne\ell).
\end{equation}
Consequently, for $j\ne\ell$,
\begin{align}\label{cs:eq:held-projection-proof}
 \mathbb E_{H,\ell}[y_{a_{\ell j}}y_{a_\ell}]
 &=\frac1{qD}\sum_{(e,r)\in H}y_{r,e}
      \underbrace{\frac1N\sum_t y_{t,e}}_{0}=0,\notag\\
 \mathbb E_{H,\ell}[\omega_{a_{\ell j}}y_{a_{\ell j}}y_{a_\ell}]
 &=\frac1{qDN}\sum_{(e,r)\in H}y_{r,e}\sum_t\omega_{t,e}y_{t,e}=Z_S.
\end{align}
For $j=\ell$, the two averages are $\nu_H$ and $0$, since
$a_\ell\in H\Rightarrow\omega_{a_\ell}=0$. Substitution in
\[
 \frac{\lambda}{L-1}\sum_{\ell=2}^L\sum_j
  (\beta K_{j\ell}+\lambda U_{j\ell})
  \mathbb E_{H,\ell}[(\beta+\lambda\omega_{a_{\ell j}})
                             y_{a_{\ell j}}y_{a_\ell}]
\]
gives~\eqref{cs:eq:PS}. In general, $Z_S\ne0$.
\end{proof}

\begin{lemma}[Concentration of the balanced cubic]\label{cs:lem:mask}
Fix $L\ge2$, $p\in(0,1)$, and $\delta\in(0,1)$. For sufficiently large
$N$, with probability at least $1-\delta$,
\begin{equation}\label{cs:eq:mask-result}
 \begin{gathered}
 F_0(S):=\Pr_{\rm full}(\chi\text{ is valid})\ge p^L/2,
 \qquad D\Pr_S(a_{\ell j}=a)\le J_0:=2p^{-L},\\
 \left\|P_S/\nu^2-g\right\|_{[2]}\le\eta_{\rm mask}
 =O_{L,p}\!\left(\frac{\sqrt{\log(2/\delta)}}N\right),
 \qquad g(\beta)=\beta\lambda\left(\beta+\frac\lambda{pL}\right).
 \end{gathered}
\end{equation}
An explicit $\eta_{\rm mask}$ is given in~\eqref{cs:eq:mask-explicit}.
\end{lemma}
\begin{proof}
\csproofstep{1. Raw moments and exact normalizations.}
Let $I_S(\chi)$ indicate validity and set
\[
 F_\ell=\mathbb E_{\rm full}[I_S(\chi)y_{a_\ell}^2],\qquad
 h_S(a)=\Pr_{\rm full}(I_S(\chi)=1\mid a_1=a).
\]
Given $\chi$, let $\chi^{(\ell)}$ start at $e_{\ell-1}$ with its first
edge forced to be $a_\ell$ and its remaining $L-1$ relations fresh.
Let $r_0$ be independent and uniform. Define
\begin{equation}\label{cs:eq:raw-definitions}
 \begin{aligned}
 T_\ell&=\mathbb E[I_S(\chi)I_S(\chi^{(\ell)})y_{a_\ell}^2],\\
 M_S&=\mathbb E[I_S(\chi)\mathbf1\{(e_0,r_0)\in H\}
                            y_{r_0,e_0}y_{r_1,e_0}].
 \end{aligned}
\end{equation}
Then
\begin{align}\label{cs:eq:raw-identities}
 \Pr_S(a_1=a)&=\frac{h_S(a)}{DF_0},&
 \omega_a&=\frac{h_S(a)}{LF_0},\notag\\
 K_{\ell\ell}&=\frac{F_\ell}{F_0},&
 U_{\ell\ell}&=\frac{\mathbb E_{\rm full}[I_S h_S(a_\ell)y_{a_\ell}^2]}{LF_0^2}
               =\frac{T_\ell}{LF_0^2},\notag\\
 M_S&=\frac1{DN}\sum_e\sum_{r:(e,r)\in H}y_{r,e}
                         \sum_t h_S(e,t)y_{t,e},&
 Z_S&=\frac{M_S}{qLF_0}.
\end{align}

\csproofstep{2. Explicit expectation-bias bounds.}
For complete shifts,
\begin{equation}\label{cs:eq:iid-vertices}
 \begin{gathered}
 (e_0,r_1,\ldots,r_L)\longleftrightarrow(e_0,e_1,\ldots,e_L),
 \qquad r_i=1+((e_i-e_{i-1})\bmod N),\\
 \Pr(e_0=x_0,\ldots,e_L=x_L)=N^{-(L+1)}.
 \end{gathered}
\end{equation}
The independent vertex samples underlying each configuration are
\[
 \begin{array}{c|c|c}
 \text{statistic}&\text{independent samples}&\text{distinct requests if no collisions}\\\hline
 F_0,F_\ell&e_0,\ldots,e_L&L\text{ retained}\\
 T_\ell&e_0,\ldots,e_L,z_2,\ldots,z_L&(2L-1)\text{ retained}\\
 M_S&e_0,\ldots,e_L,z&L\text{ retained},\ 1\text{ held}
 \end{array}
\]
The auxiliary path starts with $(e_{\ell-1},e_\ell)$; the mixed test edge
is $(e_0,z)$. For the event $\mathcal C$ that two independent samples coincide,
\begin{equation}\label{cs:eq:collision-prob}
 \Pr(\mathcal C)\le\gamma_N:=\frac{\binom{2L}{2}}N.
\end{equation}
Assume $pD\ge2L$. If $k\le2L$ distinct retained facts are requested,
\begin{align}\label{cs:eq:factorial-bound}
 p_k&:=\frac{(pD)_k}{(D)_k},&
 \frac{p_k}{p^k}
 &=\prod_{j=0}^{k-1}\frac{1-j/(pD)}{1-j/D}\le1,\notag\\
 \frac{p_k}{p^k}
 &\ge\prod_{j=0}^{k-1}\left(1-\frac{j}{pD}\right)
 \ge1-\sum_{j=0}^{k-1}\frac{j}{pD}
 =1-d_k,&d_k&:=\frac{k(k-1)}{2pD}.
\end{align}
For the collision-free mixed configuration,
\begin{equation}\label{cs:eq:mixed-inclusion}
 \frac{\Pr(\text{retained path, held test edge})}{qp^L}
 =\frac{p_L/p^L}{1-L/D},\qquad
 \left|\frac{p_L/p^L}{1-L/D}-1\right|
 \le\frac{d_L+L/D}{1-L/D}\le2d_L+2L/D.
\end{equation}
If $R$ is a normalized inclusion probability and $|w|\le1$ a scalar
weight, then
\begin{equation}\label{cs:eq:bias-split}
 |\mathbb E[wR]-\mathbb E w|
 \le\mathbb E[|R-1|\mathbf1_{\mathcal C^c}]
      +\mathbb E[|R-1|\mathbf1_{\mathcal C}].
\end{equation}
On $\mathcal C$, use respectively $0\le R\le p^{-L}$,
$p^{-(2L-1)}$, or $(qp^L)^{-1}$. Therefore, with expectation over masks,
\begin{equation}\label{cs:eq:raw-bias}
 \begin{array}{c|c|c}
 \text{statistic }X&\text{reference }x_\circ&|\mathbb E_{\rm mask}X-x_\circ|\le\\\hline
 F_0/p^L&1&b_0:=(1+p^{-L})\gamma_N+d_L\\
 F_\ell/p^L&\nu&b_0\\
 T_\ell/p^{2L-1}&\nu&b_T:=(1+p^{-(2L-1)})\gamma_N+d_{2L-1}\\
 M_S/(qp^L)&0&b_M:=(1+(qp^L)^{-1})\gamma_N+2d_L+2L/D\\
 \nu_H&\nu&0
 \end{array}
\end{equation}
The reference means use $\mathbb E\zeta_{e_\ell}^2=\nu$ and
$\mathbb E[\zeta_z\zeta_{e_1}]=\mathbb E\zeta_z\,\mathbb E\zeta_{e_1}=0$;
$z,e_1$ are independent before masking.

\csproofstep{3. Global sensitivity and concentration.}
If masks $S,S'$ differ by one exchange, $|S\triangle S'|=2$.
A configuration with $k$ edge requests can change only if one request
hits a changed fact. Each request is marginally uniform, so
\begin{equation}\label{cs:eq:exchange-count}
 |\mathbb E[wI_S]-\mathbb E[wI_{S'}]|
 \le\Pr(\exists i\le k:a_i\in S\triangle S')
 \le\sum_{i=1}^k\frac2D=\frac{2k}D.
\end{equation}
The two forced copies of a diagonal-pair edge count as one request.
Thus the global exchange constants are
\begin{equation}\label{cs:eq:global-lipschitz}
 \begin{array}{c|c}
 F_0/p^L,\ F_\ell/p^L&c_0=2L/(Dp^L)\\
 T_\ell/p^{2L-1}&c_T=2(2L-1)/(Dp^{2L-1})\\
 M_S/(qp^L)&c_M=2(L+1)/(Dqp^L)\\
 \nu_H&c_H=2/(qD).
 \end{array}
\end{equation}
For a function $X$ with exchange constant $c$, reveal the smaller subset
in $k_*=\min(p,q)D$ steps. Its Doob increments $\Delta_i$ obey
\begin{equation}\label{cs:eq:doob}
 \mathbb E[\Delta_i\mid\mathcal F_{i-1}]=0,\qquad
 \operatorname{range}(\Delta_i\mid\mathcal F_{i-1})\le c.
\end{equation}
The range bound follows by coupling two possible next reveals with one
exchange in their uniform completions. For the conditional log moment
function $\psi_i(s)$,
\begin{equation}\label{cs:eq:hoeffding-proof}
 \psi_i(0)=\psi_i'(0)=0,\quad
 \psi_i''(s)=\operatorname{Var}_s(\Delta_i\mid\mathcal F_{i-1})\le c^2/4,
 \quad \psi_i(s)\le s^2c^2/8.
\end{equation}
Iteration and exponential Markov yield
\begin{align}\label{cs:eq:slice}
 \mathbb E e^{s(X-\mathbb EX)}&\le e^{k_*s^2c^2/8},\notag\\
 \Pr(|X-\mathbb EX|>x)
 &\le2\inf_{s>0}e^{-sx+k_*s^2c^2/8}
 =2e^{-2x^2/(k_*c^2)}.
\end{align}
There are $J_*=2L+1$ statistics in~\eqref{cs:eq:raw-bias}. Set
\begin{equation}\label{cs:eq:event-errors}
 \begin{gathered}
 v_\delta=\sqrt{\frac{k_*}{2}\log\frac{2J_*}{\delta}},\\
 \eta_0=b_0+c_0v_\delta,\quad
 \eta_T=b_T+c_Tv_\delta,\quad
 \eta_M=b_M+c_Mv_\delta,\quad
 \eta_H=c_Hv_\delta.
 \end{gathered}
\end{equation}
A union bound gives an event $\mathcal E$, $\Pr(\mathcal E)\ge1-\delta$, on which
\begin{equation}\label{cs:eq:raw-event}
 \begin{gathered}
 |F_0/p^L-1|,\ |F_\ell/p^L-\nu|\le\eta_0,\qquad
 |T_\ell/p^{2L-1}-\nu|\le\eta_T,\\
 |M_S/(qp^L)|\le\eta_M,\qquad |\nu_H-\nu|\le\eta_H.
 \end{gathered}
\end{equation}

\csproofstep{4. Ratios and cubic coefficients.}
Assume $\eta_0\le1/2$ and write $z=F_0/p^L$. On $\mathcal E$,
\begin{align}\label{cs:eq:ratio-bounds}
 F_0&\ge p^L(1-\eta_0)\ge p^L/2,\notag\\
 |K_{\ell\ell}-\nu|
 &=\left|\frac{F_\ell/p^L-\nu z}{z}\right|
 \le k_\Delta:=\frac{(1+\nu)\eta_0}{1-\eta_0},\notag\\
 \left|U_{\ell\ell}-\frac\nu{pL}\right|
 &=\frac1{pL}\left|\frac{T_\ell/p^{2L-1}-\nu z^2}{z^2}\right|
 \le u_\Delta:=\frac{\eta_T+\nu(2\eta_0+\eta_0^2)}{pL(1-\eta_0)^2},\notag\\
 |Z_S|&\le z_\Delta:=\frac{\eta_M}{L(1-\eta_0)},\qquad
 0\le\omega_a\le\omega_+:=\frac1{Lp^L(1-\eta_0)},\notag\\
 |K_{j\ell}|&\le1,\qquad |U_{j\ell}|\le\omega_+.
\end{align}
Introduce scalar coefficients, not parameter matrices:
\[
 A_S=\frac{\nu_H}{L-1}\sum_{\ell=2}^L K_{\ell\ell},\quad
 B_S=\frac{\nu_H}{L-1}\sum_{\ell=2}^L U_{\ell\ell},\quad
 K_O=\frac1{L-1}\sum_{\ell=2}^L\sum_{j\ne\ell}K_{j\ell},\quad
 U_O=\frac1{L-1}\sum_{\ell=2}^L\sum_{j\ne\ell}U_{j\ell}.
\]
The product inequality $|xy-x_0y_0|\le |x_0||y-y_0|+|y_0||x-x_0|
+|x-x_0||y-y_0|$ gives
\begin{equation}\label{cs:eq:coefficient-bounds}
 \begin{aligned}
 |A_S-\nu^2|&\le a_\Delta:=\nu k_\Delta+\nu\eta_H+k_\Delta\eta_H,\\
 |B_S-\nu^2/(pL)|&\le b_\Delta:=\nu u_\Delta+\frac\nu{pL}\eta_H+u_\Delta\eta_H,\\
 |K_O|&\le L-1,\qquad |U_O|\le(L-1)\omega_+.
 \end{aligned}
\end{equation}
Now
\begin{equation}\label{cs:eq:cubic-difference}
 P_S-\nu^2g=(A_S-\nu^2)\beta^2\lambda
 +(B_S-\nu^2/(pL))\beta\lambda^2
 +Z_S(K_O\beta\lambda^2+U_O\lambda^3).
\end{equation}
Direct differentiation supplies all three norm constants:
\[
 \begin{array}{c|ccc|c}
 h&\|h\|_\infty&\|h'\|_\infty&\frac12\|h''\|_\infty&\|h\|_{[2]}\\\hline
 \beta^2\lambda,\ \beta\lambda^2&4/27&1&2&85/27\\
 \lambda^3&1&3&3&7
 \end{array}
\]
Hence an explicit mask error is
\begin{equation}\label{cs:eq:mask-explicit}
 {\eta_{\rm mask}:=\frac1{\nu^2}
  \left[\frac{85}{27}(a_\Delta+b_\Delta)
  +(L-1)z_\Delta\left(\frac{85}{27}+7\omega_+\right)\right].}
\end{equation}
For fixed $L,p$, all four errors in~\eqref{cs:eq:event-errors} are
$O_{L,p}(\sqrt{\log(2/\delta)}/N)$; equations
\eqref{cs:eq:ratio-bounds}--\eqref{cs:eq:mask-explicit} give the same rate.
Finally,
\[
 D\Pr_S(a_{\ell j}=a)
 =D\frac{\Pr_{\rm full}(a_{\ell j}=a,I_S=1)}{F_0}
 \le\frac{D\Pr_{\rm full}(a_{\ell j}=a)}{F_0}
 =F_0^{-1}\le2p^{-L}.
\]
\end{proof}

\begin{lemma}[Uniform finite-time cubic approximation]\label{cs:lem:time}
Fix a mask with $F_0\ge p^L/2$ and a horizon $\tau>0$. For sufficiently
large $N$, the joint raw flow satisfies
\begin{equation}\label{cs:eq:time}
 -\frac{2LD^2}{\nu^2\tau^3}
 \left[\mathcal E_H(\theta_\beta(\tau))-\frac{\nu_H}{2}\right]
 =\frac{P_S(\beta)}{\nu^2}+r_{\rm time}(\beta),
 \qquad \|r_{\rm time}\|_{[2]}\le\eta_{\rm time}=O_{L,p,\tau}(D^{-1}).
\end{equation}
An explicit bound is~\eqref{cs:eq:time-explicit}.
\end{lemma}
\begin{proof}
\csproofstep{1. Exact raw gradients.}
For a chain function $h$, define the weighted linear functional
\[
 \mathcal T_{\beta,\ell}[h]
 :=\beta\mathbf1\{\ell=1\}\mathbb E_{\rm at}h
       +\frac\lambda L\mathbb E_Sh.
\]
Write $G_{r,e}=w_rb_{r,e}$ and $R_\ell=f_\ell-y_{a_\ell}$. Then
\begin{equation}\label{cs:eq:raw-flow}
 \begin{aligned}
 Q_a&=\sum_{\ell,j}C_{j\ell}
           \mathcal T_{\beta,\ell}[\mathbf1\{a_{\ell j}=a\}R_\ell],\\
 \dot b_{r,e}&=-w_rQ_{r,e},\qquad
 \dot w_r=-\sum_e b_{r,e}Q_{r,e},\\
 \dot C_{j\ell}&=-\mathcal T_{\beta,\ell}[G_{a_{\ell j}}R_\ell].
 \end{aligned}
\end{equation}
For $\langle h,k\rangle_\beta:=\sum_\ell\mathcal T_{\beta,\ell}[h_\ell k_\ell]$,
\begin{align}\label{cs:eq:existence}
 \|y\|_\beta^2&=\beta\mathbb E_{\rm at}y_{a_1}^2
            +\frac\lambda L\sum_\ell\mathbb E_S y_{a_\ell}^2\le1,\notag\\
 \frac{d}{dt}(\|w\|_2^2+\|b\|_F^2+\|C\|_F^2)
 &=-6\langle f,f-y\rangle_\beta
 =-6\|f-y/2\|_\beta^2+\tfrac32\|y\|_\beta^2\le\tfrac32.
\end{align}
The polynomial ODE thus has a global real solution, with smooth
$\beta$-dependence on finite intervals. Here $y_\ell=y_{a_\ell}$.

\csproofstep{2. Rescaled equations and their operator bounds.}
Use $Y=Db$ only as an auxiliary coordinate. Define
\begin{equation}\label{cs:eq:maps}
 \begin{aligned}
 [\mathcal A_\beta(C)]_a
  &:=D\sum_{\ell,j}C_{j\ell}
       \mathcal T_{\beta,\ell}[\mathbf1\{a_{\ell j}=a\}y_{a_\ell}],\\
 [\mathcal B_\beta(H)]_{j\ell}
  &:=\mathcal T_{\beta,\ell}[H_{a_{\ell j}}y_{a_\ell}],\\
 [\mathcal U_\beta(C_1,C_2,H)]_a
  &:=D\sum_{\ell,j}(C_1)_{j\ell}
       \mathcal T_{\beta,\ell}\!\left[
        \mathbf1\{a_{\ell j}=a\}\sum_k(C_2)_{k\ell}H_{a_{\ell k}}\right],\\
 [\mathcal V_\beta(H_1,C,H_2)]_{j\ell}
  &:=\mathcal T_{\beta,\ell}\!\left[
         (H_1)_{a_{\ell j}}\sum_k C_{k\ell}(H_2)_{a_{\ell k}}\right].
 \end{aligned}
\end{equation}
The arguments $H,H_1,H_2$ in these maps are scalar tables, not held sets.
Let $J_0=2p^{-L}$; subscripts $1,0$ below denote $\beta=1,0$.
The candidate probabilities give the base operator bounds
\begin{equation}\label{cs:eq:base-operator-bounds}
 \begin{aligned}
 \|\mathcal A_1(C)\|_{\max}
 &\le D\frac1D\sum_j|C_{j1}|\le\|C\|_1,
 &\|\mathcal A_0(C)\|_{\max}&\le\frac{J_0}{L}\|C\|_1,\\
 \|\mathcal B_1(H)\|_1&\le L\|H\|_{\max},
 &\|\mathcal B_0(H)\|_1&\le\frac{L^2}{L}\|H\|_{\max},\\
 \|\mathcal U_1(C_1,C_2,H)\|_{\max}
 &\le\|C_1\|_1\|C_2\|_1\|H\|_{\max},
 &\|\mathcal U_0(C_1,C_2,H)\|_{\max}
 &\le\frac{J_0}{L}\|C_1\|_1\|C_2\|_1\|H\|_{\max},\\
 \|\mathcal V_1(H_1,C,H_2)\|_1
 &\le L\|H_1\|_{\max}\|C\|_1\|H_2\|_{\max},
 &\|\mathcal V_0(H_1,C,H_2)\|_1
 &\le\|H_1\|_{\max}\|C\|_1\|H_2\|_{\max}.
 \end{aligned}
\end{equation}
For example, the final coefficient is
$L^{-1}\sum_{j,\ell,k}|C_{k\ell}|=\|C\|_1$.
Apply~\eqref{cs:eq:product} to the affine mixture of these operators:

\begin{equation}\label{cs:eq:operator-bounds}
 \begin{aligned}
 \|\mathcal A_\beta(C)\|_{[2]}
   &\le 2(1+J_0/L)\|C\|_{[2]},\\
 \|\mathcal U_\beta(C_1,C_2,H)\|_{[2]}
   &\le2(1+J_0/L)\|C_1\|_{[2]}\|C_2\|_{[2]}\|H\|_{[2]},\\
 \|\mathcal B_\beta(H)\|_{[2]}
   &\le(2L+2L)\|H\|_{[2]},\\
 \|\mathcal V_\beta(H_1,C,H_2)\|_{[2]}
   &\le(2L+2)\|H_1\|_{[2]}\|C\|_{[2]}\|H_2\|_{[2]}.
 \end{aligned}
\end{equation}
Set
\begin{equation}\label{cs:eq:K}
 K:=4(L+1+J_0)\ge\max\{2(1+J_0/L),4L,2L+2\}.
\end{equation}
Row scaling is $(wY)_{r,e}=w_rY_{r,e}$. Substitution into the raw flow gives
\begin{equation}\label{cs:eq:exact-rescaled}
 \begin{aligned}
 \dot Y&=w\mathcal A_\beta(C)
          -D^{-1}w\mathcal U_\beta(C,C,wY),\\
 \dot w_r&=D^{-2}\sum_eY_{r,e}[\mathcal A_\beta(C)]_{r,e}
          -D^{-3}\sum_eY_{r,e}[\mathcal U_\beta(C,C,wY)]_{r,e},\\
 \dot C&=D^{-1}\mathcal B_\beta(wY)
          -D^{-2}\mathcal V_\beta(wY,C,wY).
 \end{aligned}
\end{equation}
Thus, for $s_Y=\|Y\|_{[2]}$, $s_w=\|w\|_{[2]}$, $s_C=\|C\|_{[2]}$,
\begin{equation}\label{cs:eq:bootstrap}
 \begin{aligned}
 \|\dot Y\|_{[2]}&\le K s_ws_C(1+s_ws_Cs_Y/D),\\
 \|\dot C\|_{[2]}&\le (K/D)s_ws_Y(1+s_ws_Cs_Y/D),\\
 \|\dot w\|_{[2]}&\le (KN/D^2)s_Ys_C(1+s_ws_Cs_Y/D).
 \end{aligned}
\end{equation}

\csproofstep{3. Close the bootstrap with a stated dimension threshold.}
Let $R_Y=1+8K\tau$ and assume
\begin{equation}\label{cs:eq:time-threshold}
 D\ge\max\{4R_Y,\ 8KR_Y\tau\}.
\end{equation}
Up to the first exit from $s_w,s_C\le2$, $s_Y\le R_Y$,
\begin{equation}\label{cs:eq:bootstrap-close}
 \begin{gathered}
 s_ws_Cs_Y/D\le4R_Y/D\le1,\qquad
 \|\dot Y\|_{[2]}\le8K,\\
 \|\dot C\|_{[2]}\le4KR_Y/D,\qquad
 \|\dot w\|_{[2]}\le4KNR_Y/D^2\le4KR_Y/D,\\
 s_Y(t)\le8Kt<R_Y,\qquad
 s_C(t),s_w(t)\le1+4KR_Y\tau/D\le\tfrac32<2.
 \end{gathered}
\end{equation}
No exit occurs on $[0,\tau]$. Substituting $s_Y(s)\le8Ks$ back into
\eqref{cs:eq:bootstrap} sharpens the other two estimates:
\begin{equation}\label{cs:eq:rough}
 \begin{aligned}
 \|Y(t)\|_{[2]}&\le8Kt,\\
 \|C(t)-E_{11}\|_{[2]}&\le\frac{4K}{D}\int_0^t8Ks\,ds
                                  =\frac{16K^2t^2}{D},\\
 \|w(t)-\mathbf1\|_{[2]}&\le\frac{4KN}{D^2}\int_0^t8Ks\,ds
                                  =\frac{16K^2Nt^2}{D^2}.
 \end{aligned}
\end{equation}

\csproofstep{4. Derive the first two expansions by integration.}
Let
\begin{equation}\label{cs:eq:v}
 v_a:=[\beta+\lambda\omega_a]y_a
       =[\mathcal A_\beta(E_{11})]_a,
 \qquad \|v\|_{[2]}\le K.
\end{equation}
Using $N\le D$, equations~\eqref{cs:eq:exact-rescaled} and
\eqref{cs:eq:rough} imply
\begin{align}\label{cs:eq:Y-error}
 \|\dot Y-v\|_{[2]}
 &\le\|w-\mathbf1\|_{[2]}K s_C
       +K\|C-E_{11}\|_{[2]}+\frac KDs_w^2s_C^2s_Y\notag\\
 &\le\frac{32K^3Nt^2}{D^2}+\frac{16K^3t^2}{D}
                        +\frac{128K^2t}{D}
 \le\frac{128K^2t+48K^3t^2}{D},\notag\\
 \|Y(t)-tv\|_{[2]}
 &\le\frac{64K^2t^2+16K^3t^3}{D}.
\end{align}
Consequently,
\begin{align}\label{cs:eq:G-error}
 \|wY-tv\|_{[2]}
 &\le\|Y-tv\|_{[2]}+\|w-\mathbf1\|_{[2]}\|Y\|_{[2]}\notag\\
 &\le\frac{64K^2t^2+16K^3t^3}{D}
        +\frac{128K^3Nt^3}{D^2}
 \le\frac{E_G(t)}D,\qquad
 E_G(t):=64K^2t^2+144K^3t^3.
\end{align}
For routing, the exact equation gives
\begin{align}\label{cs:eq:C-error}
 \left\|\dot C-\frac tD\mathcal B_\beta(v)\right\|_{[2]}
 &\le\frac KD\|wY-tv\|_{[2]}
       +\frac K{D^2}s_w^2s_Y^2s_C\notag\\
 &\le\frac{K E_G(t)+512K^3t^2}{D^2},\notag\\
 \left\|C-E_{11}-\frac{t^2}{2D}\mathcal B_\beta(v)\right\|_{[2]}
 &\le\frac1{D^2}\int_0^t[K E_G(s)+512K^3s^2]ds
 =\frac{E_C(t)}{D^2},\notag\\
 E_C(t)&:=192K^3t^3+36K^4t^4.
\end{align}
In particular, the last two bounds are
\begin{equation}\label{cs:eq:jets}
 \left\|G-\frac tDv\right\|_{[2]}\le\frac{E_G(t)}{D^2},\qquad
 \left\|C-E_{11}-\frac{t^2}{2D}\mathcal B_\beta(v)\right\|_{[2]}
 \le\frac{192K^3t^3+36K^4t^4}{D^2}.
\end{equation}

\csproofstep{5. Multiply the expansions and project onto the target.}
Let $C^{(2)}=t^2\mathcal B_\beta(v)/(2D)$ and $G^{(1)}=tv/D$.
For $\ell\ge2$, $C_{:\ell}(0)=0$ and
\[
 [\mathcal B_\beta(v)]_{j\ell}
  =\frac\lambda L[\beta K_{j\ell}+\lambda U_{j\ell}].
\]
The cubic prediction is therefore
\begin{equation}\label{cs:eq:f3}
 f_\ell^{(3)}(t):=\sum_j C^{(2)}_{j\ell}G^{(1)}_{a_{\ell j}}
  =\frac{t^3\lambda}{2LD^2}
       \sum_j(\beta K_{j\ell}+\lambda U_{j\ell})v_{a_{\ell j}}.
\end{equation}
The product remainder satisfies
\begin{align}\label{cs:eq:f-error}
 \|f_\ell-f_\ell^{(3)}\|_{[2]}
 &\le\|C_{:\ell}-C^{(2)}_{:\ell}\|_{[2]}\|G\|_{[2]}
       +\|C^{(2)}_{:\ell}\|_{[2]}\|G-G^{(1)}\|_{[2]}\notag\\
 &\le\frac{E_C(t)}{D^2}\frac{16Kt}D
       +\frac{K^2t^2}{2D}\frac{E_G(t)}{D^2}
 =\frac{E_f(t)}{D^3},\notag\\
 E_f(t)&:=16KtE_C(t)+\tfrac12K^2t^2E_G(t)
        =3104K^4t^4+648K^5t^5,\notag\\
 \|f_\ell(t)\|_{[2]}
 &\le\frac{16K^2t^2}D\frac{16Kt}D
 =\frac{F(t)}{D^2},\qquad F(t):=256K^3t^3.
\end{align}
Lemma~\ref{cs:lem:balance} supplies the missing target-projection step:
\begin{align}\label{cs:eq:cubic-projection}
 \mathbb E_{H,\ell}[v_{a_{\ell j}}y_{a_\ell}]
 &=\beta\nu_H\mathbf1\{j=\ell\}+\lambda Z_S\mathbf1\{j\ne\ell\},\notag\\
 \frac1{L-1}\sum_{\ell=2}^L\mathbb E_{H,\ell}[y_{a_\ell}f_\ell^{(3)}]
 &=\frac{t^3\lambda}{2LD^2(L-1)}\sum_{\ell=2}^L\sum_j
     (\beta K_{j\ell}+\lambda U_{j\ell})
     (\beta\nu_H\mathbf1\{j=\ell\}+\lambda Z_S\mathbf1\{j\ne\ell\})\notag\\
 &=\frac{t^3}{2LD^2}P_S(\beta).
\end{align}
Finally,
\begin{align}\label{cs:eq:loss-remainder}
 \mathcal E_H-\frac{\nu_H}{2}+\frac{t^3P_S}{2LD^2}
 &=\frac1{L-1}\sum_{\ell=2}^L\mathbb E_{H,\ell}
          [-y_{a_\ell}(f_\ell-f_\ell^{(3)})+f_\ell^2/2],\notag\\
 \left\|\mathcal E_H-\frac{\nu_H}{2}+\frac{t^3P_S}{2LD^2}\right\|_{[2]}
 &\le\frac{E_f(t)}{D^3}+\frac{F(t)^2}{2D^4}.
\end{align}
Thus an explicit time error is
\begin{equation}\label{cs:eq:time-explicit}
 \begin{aligned}
 \eta_{\rm time}&:=\frac{2L}{\nu^2\tau^3}
       \left[\frac{E_f(\tau)}D+\frac{F(\tau)^2}{2D^2}\right]\\
 &={\frac{2L}{\nu^2}\left[
       \frac{3104K^4\tau+648K^5\tau^2}{D}
        +\frac{32768K^6\tau^3}{D^2}\right]}
   =O_{L,p,\tau}(D^{-1}).
 \end{aligned}
\end{equation}
All bounds include two derivatives in $\beta$ by~\eqref{cs:eq:product}.
\end{proof}

\begin{lemma}[Explicit center and stable ordering]\label{cs:lem:optimizer}
Set $a=(pL)^{-1}$, $g(\beta)=\beta(1-\beta)[\beta+a(1-\beta)]$,
$\mu=2\min\{a,1\}$, and $m=\min\{a,1\}/3$.
If $J=g+r$ and $\|r\|_{[2]}\le\eta<m$, then $J$ has a unique
maximizer $\beta^*\in(1/3,2/3)$ and
\begin{equation}\label{cs:eq:stability}
 |\beta^*-\beta(pL)|\le\frac{\eta}{\mu-2\eta},\qquad
 \beta'(z)>0\quad(z>0).
\end{equation}
\end{lemma}
\begin{proof}
Direct differentiation gives
\begin{equation}\label{cs:eq:reference-derivatives}
 \begin{gathered}
 g'=a+2(1-2a)\beta+3(a-1)\beta^2,\qquad
 g''=2(1-2a)+6(a-1)\beta,\\
 g'(0)=a,\quad g'(1/3)=1/3,\quad g'(2/3)=-a/3,\quad g'(1)=-1,\\
 -g''(1/3)=2a,\quad -g''(2/3)=2,\quad
 \inf_{[1/3,2/3]}(-g'')=\mu.
 \end{gathered}
\end{equation}
A quadratic with opposite endpoint signs has exactly one root in $(0,1)$.
Writing $b\in[0,1)$ for a candidate mixture and $x=b/(1-b)$,
\begin{equation}\label{cs:eq:center-solve}
 g'(b)=0\iff-x^2+2(1-a)x+a=0
 \iff x=1-a+\sqrt{1-a+a^2}
 \iff b=\beta(pL)\in(1/3,2/3).
\end{equation}
For the exterior slopes,
\begin{equation}\label{cs:eq:exterior-slopes}
 \begin{aligned}
 \inf_{[0,1/3]}g'&\ge
  \begin{cases}\min\{a,1/3\},&a\le1,\\1/3,&a\ge1,\end{cases}
  \quad\ge m,\\
 \inf_{[2/3,1]}(-g')&\ge
  \begin{cases}a/3,&a\le1,\\\min\{a/3,1\},&a\ge1,\end{cases}
  \quad\ge m.
 \end{aligned}
\end{equation}
These follow from the sign of $g'''=6(a-1)$ and the endpoint values
in~\eqref{cs:eq:reference-derivatives}. Hence
\begin{equation}\label{cs:eq:localize}
 \begin{gathered}
 J'\ge m-\eta>0\quad\text{on }[0,1/3],\qquad
 J'\le-m+\eta<0\quad\text{on }[2/3,1],\\
 -J''\ge\mu-2\eta>0\quad\text{on }[1/3,2/3].
 \end{gathered}
\end{equation}
Existence follows by compactness; these inequalities give interiority and uniqueness.
Writing $\theta=\beta(pL)$, we have
\begin{equation}\label{cs:eq:optimizer-displacement}
 (\mu-2\eta)|\beta^*-\theta|
 \le|J'(\beta^*)-J'(\theta)|=|r'(\theta)|\le\eta.
\end{equation}
For $\Delta=z^2-z+1$,
\begin{equation}\label{cs:eq:center-trend}
 \beta'(z)=\beta(z)^2
       \left(1-\frac{2z-1}{2\sqrt\Delta}\right)>0,
 \qquad 4\Delta-(2z-1)^2=3.
\end{equation}
\end{proof}

\begin{proof}[Proof of Theorem~\ref{cs:thm:main}]
On the event of Lemma~\ref{cs:lem:mask}, let
\begin{equation}\label{cs:eq:total-error}
 \begin{aligned}
 J_\tau(\beta)&:=-\frac{2LD^2}{\nu^2\tau^3}
       [\mathcal E_H(\theta_\beta(\tau))-\nu_H/2]=g(\beta)+r(\beta),\\
 \|r\|_{[2]}&\le\eta_N:=\eta_{\rm mask}+\eta_{\rm time}
   =O_{L,p,\tau}\!\left(\frac{\sqrt{\log(2/\delta)}}N+\frac1{N^2}\right).
 \end{aligned}
\end{equation}
The explicit sufficient conditions are
\begin{equation}\label{cs:eq:sufficient}
 pN^2\ge2L,\qquad \eta_0\le\tfrac12,\qquad
 N^2\ge\max\{4R_Y,8KR_Y\tau\},\qquad \eta_N<m.
\end{equation}
They hold for sufficiently large $N$ with $L,p,\tau,\delta$ fixed.
Lemma~\ref{cs:lem:optimizer} gives
\begin{equation}\label{cs:eq:final-radius}
 {\mathfrak R_N:=\frac{\eta_N}{\mu-2\eta_N}
   =O_{L,p,\tau}\!\left(\frac{\sqrt{\log(2/\delta)}}N\right),\qquad
 |\beta^*_{\tau,S}-\beta(pL)|\le\mathfrak R_N.}
\end{equation}
Maximizing $J_\tau$ is equivalent to minimizing $\mathcal E_H$. Moreover,
\begin{equation}\label{cs:eq:trend-transfer}
 \begin{gathered}
 \partial_L\beta(pL)=p\beta'(pL)>0,\qquad
 \partial_q\beta(L(1-q))=-L\beta'(pL)<0.
 \end{gathered}
\end{equation}
\end{proof}

\rebuttal{
\section{Additional details of experiments}

\subsection{Additional details of \Cref{fig:toy}}
\label{app:toy}

Atomic examples use the same prompt format as reasoning
examples but supervise only the first transition. We sample the
starting state uniformly from all states and the first relation
uniformly and independently from all relations. Thus, the supervised
state--relation pair is uniform over the entire fact table, including
pairs held out from reasoning supervision. After the first transition,
each subsequent relation is sampled uniformly from the seen outgoing
relations of the current state, and the state is updated accordingly.
Cross-entropy is applied only to the first successor state; all
remaining targets are masked. All runs use $4000$ AdamW updates with learning rate $0.002$,
weight decay $0.01$, and batch size $256$.

\begin{figure}[t]
    \centering
    \includegraphics[width=\linewidth]{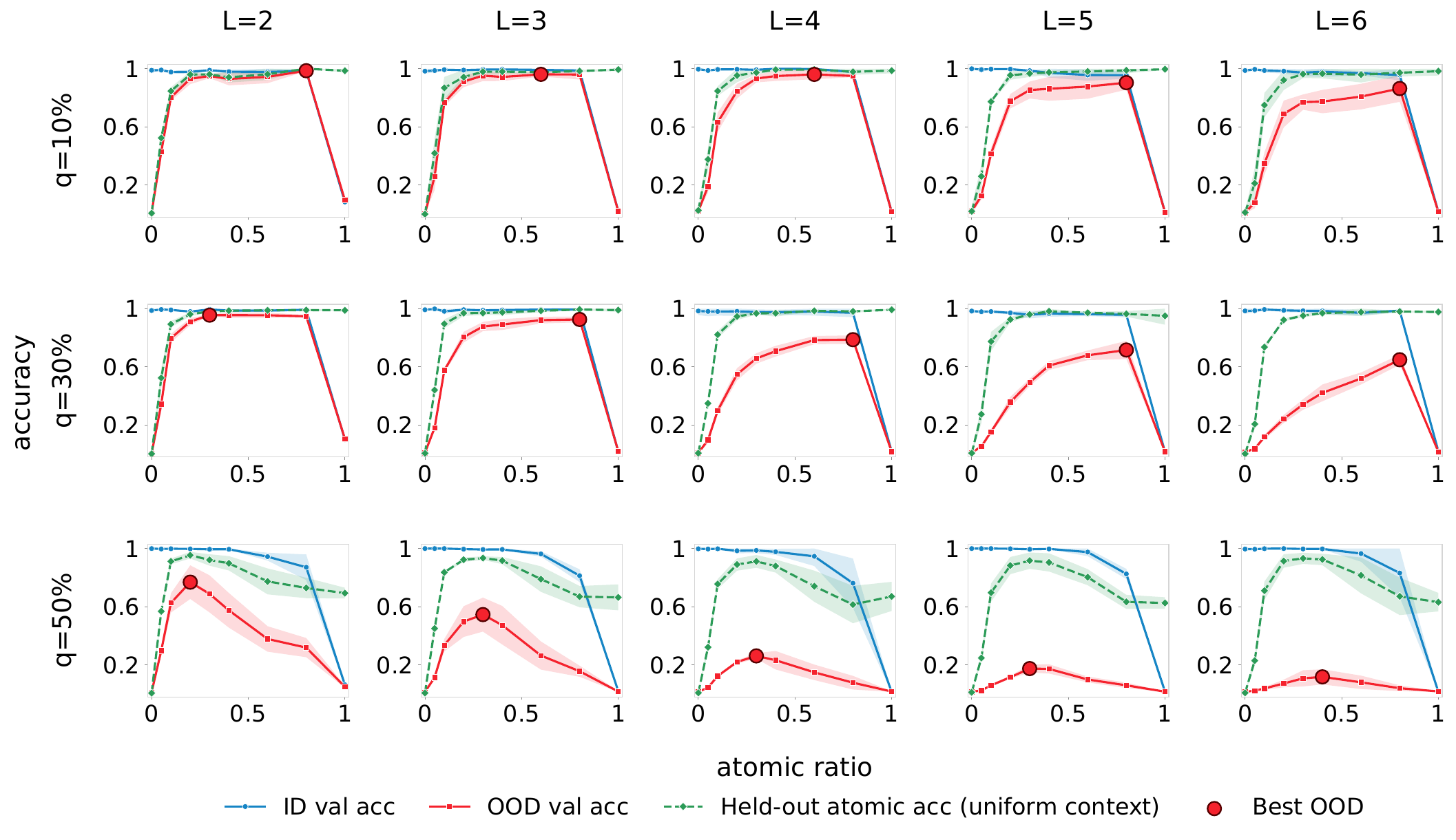}
    \caption{\textbf{Held-out atomic recall.}
    \rebuttal{ID and OOD final-answer accuracy and held-out atomic accuracy versus atomic example ratio, using the same setting as \Cref{fig:toy}. For held-out atomic evaluation (green dashed), we sample \((e_0,r_1)\) uniformly from held-out facts and \(r_2,\ldots,r_L\) independently and uniformly from all relations, scoring only \(e_1=\pi_{r_1}(e_0)\). Curves and shaded regions show the mean and ±1 standard deviation across three training seeds.}
}
    \label{fig:toy_with_atom_acc}
\end{figure}

\subsection{Additional details of \Cref{fig:multihop-mixture}}
\label{app:multihop}

\paragraph{Knowledge world and reasoning examples.}
We construct a synthetic world containing 32 people and 32 relation rules, giving 1,024 atomic facts.  An atomic document states one person--relation fact. An $H$-hop reasoning example specifies a starting person and $H$ relations, followed by a worked solution listing the intermediate transitions and final person.

\paragraph{Examples of training data.}
The following excerpts are taken directly from the generated corpora. The atomic-pretraining examples are:
\begin{quote}
\small
\textbf{Atomic document 1:} Nevaeh Tristan Kay's assessor is Jesse Alexis McCann.

\textbf{Atomic document 2:} Jesse Alexis McCann's trainer is Delaney Emma Chadwick.
\end{quote}
The corresponding SFT-format reasoning example composes these two facts:
\begin{quote}
\small
\textbf{Question:} Starting from Nevaeh Tristan Kay, who do you reach after following: assessor, trainer?

\textbf{Answer:} Nevaeh Tristan Kay's assessor is Jesse Alexis McCann. Jesse Alexis McCann's trainer is Delaney Emma Chadwick. Answer: Delaney Emma Chadwick.
\end{quote}

\paragraph{Reasoning holdout and data splits.}
For each holdout fraction $q$, we reserve a subset of person--relation pairs that may not occur in reasoning-training chains. These subsets are nested across $q$, and every reserved fact remains in the atomic corpus. Training chains and ID evaluation queries use only unreserved pairs; OOD queries contain at least one reserved pair. Each setting contains 3,000 reasoning-training rows, 300 ID queries, and 300 OOD queries. At $H=2,q=70\%$, only 2,921 distinct legal training chains exist, so some training rows repeat.

\paragraph{Model and training.}
We train a Transformer from scratch with 6 layers, hidden dimension 384, 6 attention heads (12M), and context length 512 for 300 steps. Each update contains 32,768 tokens (64 sequences of length 512). We use Muon for Transformer weight matrices with a peak learning rate of $0.02$, and AdamW for the remaining parameters.

}

\end{document}